\documentclass[11pt,a4paper]{article}
\usepackage[hyperref]{emnlp-ijcnlp-2019}
\usepackage{times}
\usepackage{latexsym}
\usepackage{graphicx}
\usepackage{graphics, amsmath}
\usepackage{amsfonts}
\usepackage{amsthm}
\usepackage{amssymb}
\usepackage{multirow}
\usepackage{enumitem}
\usepackage{makecell}
\usepackage{fixltx2e}
\usepackage{comment}
\usepackage{url}
\aclfinalcopy 

\title{Relation Module for Non-answerable Prediction \\ on Reading Comprehension}
\author{Kevin Huang, Yun Tang, Jing Huang, Xiaodong He, and Bowen Zhou \\
  JD AI Research, Mountain View, CA \\
 \tt{\{kevin.huang3, yun.tang, jing.huang} \\
 \tt{xiaodong.he, bowen.zhou \}} @jd.com}

\date{}
\begin{document}
\maketitle

\begin{abstract}
Machine reading comprehension (MRC) has attracted significant amounts of research attention recently, due to an increase of challenging reading comprehension datasets. In this paper, we aim to improve a MRC model's ability to determine whether a question has an answer in a  given context (e.g. the recently proposed SQuAD 2.0 task). 
Our solution is a relation module that is adaptable to any MRC model.
The relation module consists of both semantic extraction and relational information. We first extract high level semantics as objects from both question and context with multi-head self-attentive pooling. These semantic objects are then passed to a relation network, which generates relationship scores for each object pair in a sentence. These scores are used to determine whether a question is non-answerable.
We test the relation module on the SQuAD 2.0 dataset using both the BiDAF and BERT models as baseline readers. We obtain $1.8\%$ gain of F1 accuracy on top of the BiDAF reader, and $1.0\%$ on top of the BERT base model. These results show the effectiveness of our relation module on MRC.
\end{abstract}

\section{Introduction}
Ever since the release of many challenging large scale datasets for machine reading comprehension (MRC)~\cite{RajpurkarZLL16,JoshiCWZ17triviaqa,TrischlerWYHSBS16NewsQA,Yang18hotpotqa,CoQA,JiaL17}, there have been correspondingly many models for these datasets \cite{adamsweiyuQANet,SeoKFH16BiDAF,xiaodongliu2017san,HuPQ17nmonic,caimingxiongDCN+,Wang2018ACL,Liu2018ACL,Tay2018NIPS}. Knowing what you don't know \cite{RajpurkarJL18SQuAD2} is important in real applications of reading comprehension. Unanswerable questions are commonplace in the real world, and SQuAD 2.0 was released specifically to target this problem (see Figure~\ref{fig:exp0} for an example of non-answerable questions).

\begin{figure}[tp]
\framebox{
\parbox{0.45\textwidth}{
\small
\textbf{\textcolor{blue}{Example 1} } \newline
\textbf{Context:}  Each year, the southern California area has about 10,000 earthquakes. Nearly all of them are so small that they are not felt. Only several hundred are greater than magnitude 3.0, and only about 15–20 are greater than magnitude 4.0. The magnitude 6.7 1994 \textbf{\textcolor{red}{Northridge earthquake}} was particularly destructive, causing a substantial number of deaths, injuries, and structural collapses. It caused the most property damage of any earthquake in U.S. history, estimated at over \$20 billion.\newline
\textbf{Question:} What earthquake caused \$20 million in damage? \newline
\textbf{Answer:} None.}
}
\caption{An example of non-answerable question in SQuAD 2.0. Highlighted words are the output from the BERT base model. The true answer is ``None''.} 
\label{fig:exp0}
\end{figure}

One problem that most of the early MRC readers have in common is the inability to predict non-answerable questions. Readers on the popular SQuAD dataset have to be modified in order to accommodate a non-answerable possibility. Current methods on SQuAD 2.0 generally attempt to learn a single fully connected layer \cite{Simple_and_effective,Liu2018SAN,devlin2018bert} in order to determine whether a question/context pair is answerable. This leaves out relational information that may be useful for determining answerability.
We believe that relationships between different high-level semantics in the context are helpful to make better answerable or unanswerable decision.
For example, ``Northridge earthquake'' is mistakenly taken as the answer to the question about what earthquake caused \$20 million in damage. Because ``\$20 billon'' is positioned far away from ``Northridge earthquake'', it is hard for a model to link these two concepts together and recognize the mismatch of ``\$20 million'' in the question and ``\$20 billion'' in the context.


Motivated by exploiting high level semantic relationships in the context, our first step is to extract meaningful high-level semantics from question/context. Multi-head self-attentive pooling \cite{lin2017structured} has shown to be able to extract different views of a sentence with multiple heads.
Each head from the multi-head self attentive pooling has different weights on the context with learned parameters. This allows each head to act as a filter in order to emphasize part of the context. By summing up the weighted context, we obtain a vector representing an instance of a high-level semantic, which we can call it an ``object''. With multiple heads, we generate different semantic objects, which are then fed in to a relation network.

Relation networks \cite{Santoro2017NIPS} are specifically designed to model relationships between pairs of objects. In the case of reading comprehension, an object would ideally be phrase level semantics within a sentence. Relation networks are able to accomplish modeling these relationships by constraining the network to learn a score for each pair of these objects. After learning all of the pairwise scores, the relation network then summarizes all of the relations to a single vector. 
By taking a weighted sum of all of the relation scores that the sentence has, we generate a non-answerable score that is trained jointly with answer span scores from any MRC model to determine non-answerability.

In addition, we add in plausible answers from unanswerable examples to help train the relation module. These plausible answers help the base model learn a better span prediction and are also used to help guide our object extractor to extract relevant semantics.
We train a separate layer for start-end probabilities based on the plausible answers. We then augment the context vector with hidden states from this layer. This allows the multi-head self-attentive pooling to focus on objects related to the proposed answer span, and differentiate from other objects that are not as relevant in the context. 

In summary we propose a new relation module dedicated to learning relationships between high-level semantics and deciding whether a question is answerable. Our contributions are four-fold: 
\begin{enumerate}
\item Introduce the concept of using multi-head self-attentive pooling outputs as high level semantic objects.

\item Exploit relation networks to model the relationships between different objects in a context. We then summarize these relationships to get a final decision.

\item Introduce a separate feed-forward layer trained on plausible answers so that we can augment the context vector passed into the object extractor. This results in the object extractor extracting phrases more relevant to the proposed answer span.

\item Combining all of the above into a flexible relation module that can be added to the end of a question answering model to boost non-answerable prediction.

\end{enumerate}

To our knowledge, this is the first case of utilizing an object extractor to extract high level semantics, and a relation networks to encode relationships between these semantics in reading comprehension. Our results show improvement on top of the baseline BiDAF model and the state-of-the-art reader based on BERT, on the SQuAD 2.0 task. 

\begin{figure*}[t]
    \centering
    \includegraphics[width=2\columnwidth]{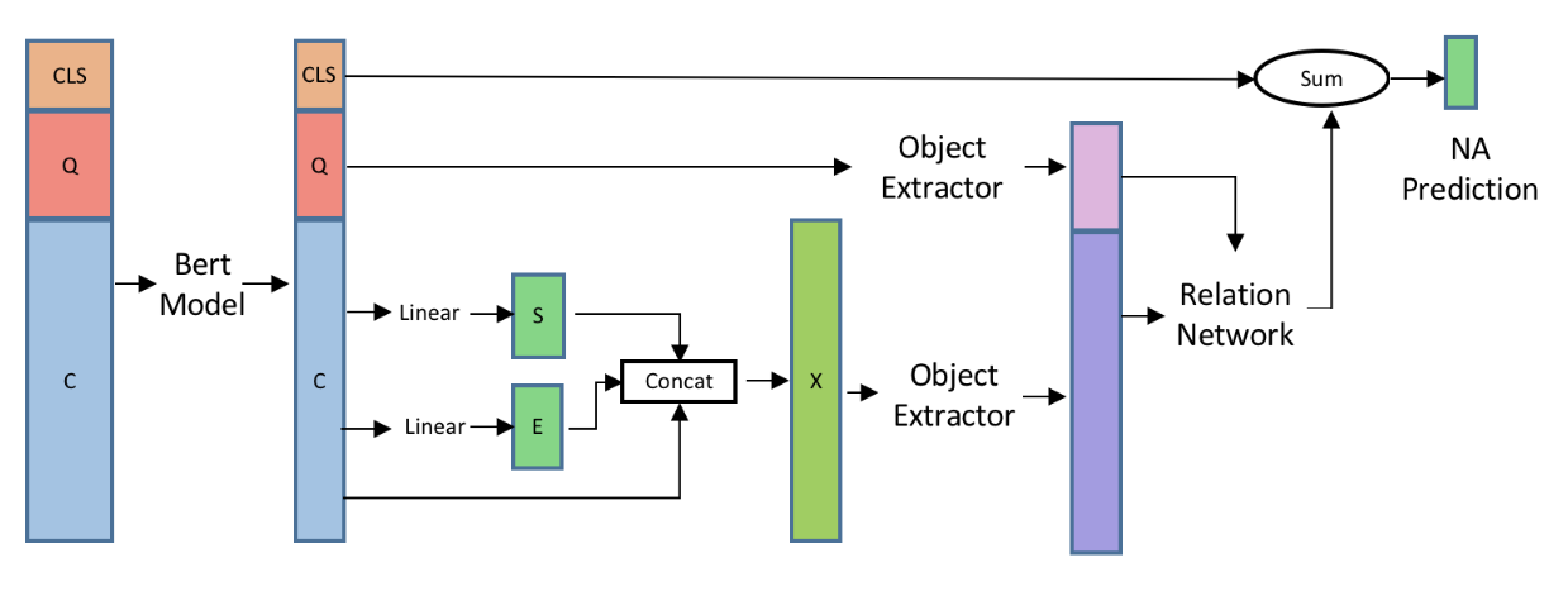}
    \caption{Relation Module on BERT. $S$ and $E$ are hidden states trained by plausible answers. We then concatenate $S$ and $E$ with the contextual representation to feed into the object extractor. After we obtain the extracted objects, we then feed into a Relation Network and pass it down for NA predictions.}
    \label{fig:RelationModule}
\label{figure:RelationModule}
\end{figure*}

\section{Related Work}
Relation Networks (RN) were first proposed by \cite{Santoro2017NIPS} in order to help neural models to reason over the relationships between two objects. Relation networks learn relationships between objects by learning a pairwise score for each object pair. Relation networks have been applied to CLEVR \cite{JohnsonHMFZG16clevr} as well as bAbI \cite{WestonBCM15babi}. 
In the CLEVR dataset, the object inputs to the relation network are visual objects in an image, extracted by a CNN, and in bAbI the object inputs are sentence encodings. In both tasks, the relation network is then used to compute a relationship score over these objects. Relation Networks were further applied to general reasoning by training the model on images \cite{HaoxuanPVRNet}. 

MAC (Memory, Attention and Composition) networks \cite{MACNetworks} are different models that have also been shown to learn relations from the CLEVR dataset. MAC networks operate with read and write cells. Each cell would compute a relation score between a knowledge base and question and write it into memory. Multiple read and write cells are strung together sequentially in order to model long chains of multi-hop reasoning. 
Although MAC networks do not explicitly reason between pairwise objects as relation networks do, MAC networks are an interesting way of generating multi-hop reasoning between objects within a context.

Another similar line of work investigated pre-training relationship embeddings across word pairs on large unlabelled corpus \cite{Jameel2018UnsupervisedLO,pair2vec}. 
These pre-trained pairwise relational embeddings were added to the attention layers of BiDAF, where higher level abstract reasoning occurs. The paper showed an impressive gain of $2.7\%$ on the SQuAD 2.0 development set on top of their version of BiDAF.

Many MRC models have been adapted to work on SQuAD 2.0 recently \cite{Hu2018ReadV,Liu2018SAN,Sun2018Unet,devlin2018bert}. \cite{Hu2018ReadV} added a separately trained answer verifier for no-answer detection with their Mnemonic Reader. The answer sentence that is proposed by the reader and the question are passed to three combinations of differently configured verifiers for fine-grained local entailment recognition. \cite{Liu2018SAN} just added one layer as the unanswerable binary classifier to their SAN reader.
\cite{Sun2018Unet} proposed the U-net with a universal node that encodes the fused information from both the question and passage. The summary U-node, question vector and two context vectors are passed to predict whether the question is answerable. Plausible answers were used for no-answer pointer prediction, while in our approach, plausible answers were used to augment context vector for object extraction that later help the no-answer prediction.

Pretraining embeddings on large unlabelled corpus has been shown to improve many downstream tasks \cite{elmoPeter2018,HowardULMFit,radfordPretrainedTranformer}. The recently released BERT \cite{devlin2018bert} greatly increased the F1 scores on the SQuAD 2.0 leaderboard. BERT consists of stacked Transformers \cite{Vaswani17transformers}, that are pre-trained on vast amounts of unlabeled data with a masked language model. The masked language model helps finetuning on downstream tasks, such as SQuAD 2.0. BERT models contains a special CLS token which is helpful for the SQuAD 2.0 task. This CLS token is trained to predict if a pair of sentences follow each other during the pre-training, which helps encode entailment information between the sentence pair. Due to a strong masked language model to help predict answers and a strong CLS token to encode entailment, BERT models are the current state-of-the art for SQuAD 2.0. 

\section{Relation Module}
Our relation module is flexible, and can be placed on top of any MRC model. We now describe the relation module in detail.
\subsection{Augmenting Inputs}
Figure~\ref{figure:RelationModule} shows our relation module on top of the base reader BERT. In addition to the original start-end prediction layers trained from true answers in the base reader, we include a separate start-end prediction layer, with separate parameters, trained specifically on plausible and true answers available in SQuAD 2.0. The context output $C$ from BERT  is projected into two hidden state layers $S$ and $E$, where $C$, $S$ and $E\in \mathbf{R}^{L\times h}$, $L$ is the context length and $h$ is the hidden size. The $S$ and $E$ layers are then projected down to a hidden dimension of 1, and trained with Cross-Entropy Loss against the plausible and true answer starts and ends. The hidden states $S$ and $E$ of this layer are concatenated with the last context layer output $C$ and projected back to the original dimension to obtain the augmented context vector $X$, which is fused with start-end span information.
\begin{align}
    S &= tanh(CW_{1}+b_{1}) \\
    E &= tanh(CW_{2}+b_{2}) \\
    X &= [C;S;E]W
\end{align}
where [;;] is concatenation of multiple tensors and $X\in \mathbf{R}^{L \times h}$.
This process is shown in Figure \ref{fig:RelationModule}, where $S$ and $E$ are hidden states trained on plausible and true answer spans. This tensor $X$ and the last question layer output $Q$ are passed to the object extractor layer.

\subsection{Object Extractor}
The augmented context tensor $X$ (and separately, question tensor $Q$) is passed through the object extractor to generate object representations from the tensor. We pass the inputs through a multi-head self-attentive pooling layer. This object extractor can be thought of as a set filters extracting out areas of interest within a sentence. We multiply the input tensor $X$ with a multi-head self attention matrix $A$ which is defined as 
\begin{align}
A &= Softmax(W_{4}\sigma(W_{3}X^{T})) \\
O &= AX
\end{align}
where $W_{3} \in R^{h \times h}$, and $W_{4} \in R^{n \times h}$;  $\sigma$ is an activation function, such as $tanh$;  $n$ is the number of heads, and $h$ is the hidden dimension. The output $O \in R^{n \times h}$ contains the $n$ objects with hidden dimension $h$ that are passed to the next layer.

\subsubsection{Object Extraction Regularization}
In order to help encourage the multiple heads to extract different meaningful semantics in the text, a regularization loss \cite{Xia2018ZeroshotUI} is introduced to encourage each head to attend to slightly different sections of the context. 
Overlapping objects centered on the answer span are expected, due to information fused from $S$ and $E$, but we do not want the entire weight distribution of the head to be solely focused on the answer span. As we show in later figures, many heads heavily weight the answer span, but also weight information relevant to the answer span needed to make a better non-answerable prediction. 
Our regularization term also helps prevent the multi-headed attentive pooling from learning a noisy distribution over all of the context. This regularization loss is defined as 
\begin{align}
    L_{aux} =\alpha || AA^{T} - I||_{2}
\end{align}
where $A$ is the weight matrix for the attention heads and $I$ is the identity matrix. $\alpha$ is set to be 0.0005 in our experiments. 

\begin{figure}
    \centering
    \includegraphics[width=1\columnwidth]{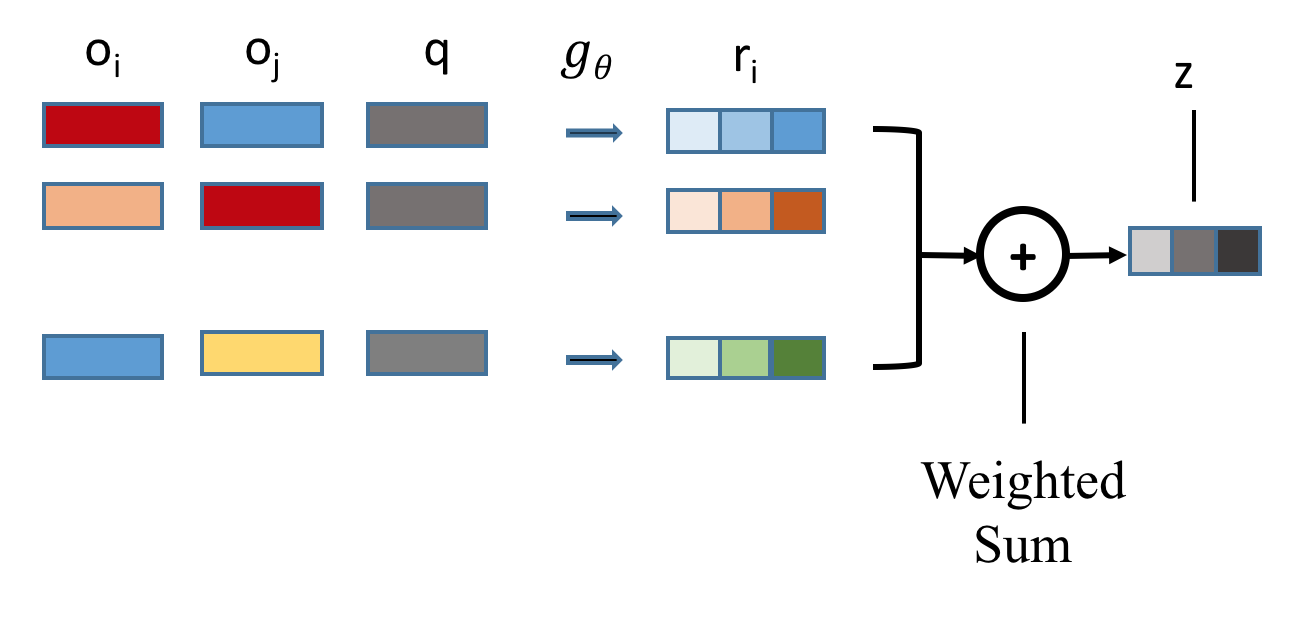}
    \caption{Illustration of a Relation Network. The $g_{\theta}$ is a MLP to score relationships between pairs}
    \label{fig:RelationNetwork}
\label{figure:RelationNetwork}
\end{figure}

\subsection{Relation Networks}
Extracted objects are subsequently passed to a relation network. We use two layer MLP $g_\theta$ (in Figure~\ref{fig:RelationNetwork}) as a scoring function to compute the similarity between objects. In the question-answering task, the context contains the contextual information necessary to determine whether a question is answerable. Phrases and ideas from various parts of the context need to come together in order to fully understand whether or not a question is answerable. Therefore our relation module takes all pairs of context objects to score, and use the question objects to guide the scoring function. We use 2 question heads $q_{0}$, $q_{1}$, so our scoring function is:

\begin{align}
r_{i} &= \sum_{j=0}^{n} \omega_{i,j} * g_{\theta}( o_{i}, o_{j}, q_{0},q_{1}) \\
z &= \sum_{i=0}^{n} \gamma_{i}*f_{\phi}(r_{i})
\end{align}
where the outputs $r_{i}$ is the weighted sum of the relation values for object $o_{i}$ from $O$, and $z$ is a summarized relation vector.  The weights $\Omega_i=[\omega_{i,0},...,\omega_{i,n}]$ and $\Gamma=[\gamma_0,...,\gamma_n]$ are computed by projecting down the relations scores into a hidden size of 1, and applying softmax. 
\begin{align}
\Omega_i &= Softmax(g_{\theta}(o_i,:,q_0,q_1)w_g )\\
\Gamma &= Softmax(f_{\phi}(:)w_f)
\end{align}
$g_{\theta}$ and $f_{\phi}$ are two layer MLP with activation function $tanh$ to compute and aggregate relational scores. Figure~\ref{figure:RelationNetwork} shows the process of a single relation network, where two context objects and question objects are passed in to $g_{\theta}$ to obtain the output $z$. 

We project the weighted sum of $f_{\phi}$ with a linear layer to a single value as a representation of the non-answerable score. This score is combined with the start/end logits from the base reader, and trained jointly with the reader's cross-entropy loss. 
By training jointly, the model is able to make a better prediction based on the confidence of the span prediction, as well as the confidence based on the non-answerable score from the relation module.

\section{Question Answering Baselines}
We test the relation module on top of our own PyTorch implementation of the BiDAF model \cite{SeoKFH16BiDAF}, as well as the recent released BERT base model \cite{devlin2018bert} for the SQuAD 2.0 task. For both of these models, we obtain improvement from adding the relation module. Note that, we do not test our relation module on top of the current leaderboard, as the details are not yet out. We also do not test on top of BERT + Synthetic Self Training \cite{BertSynthetic} due to lack of computational resources available. We are showing the effectiveness of our method and not trying to compete with the top of the leaderboard. 

\subsection{BiDAF}
We implement the baseline BiDAF model for SQuAD 2.0 task \cite{Simple_and_effective} with some modifications: adding features that are commonly used in question answering tasks such as TF-IDF, POS/NER tagging, etc, and the auxiliary training losses from \cite{Hu2018ReadV}. These modifications to the original BiDAF bring about $3.8\%$ gain of F1 on the SQuAD 2.0 development set (see Table~\ref{table:res}). 

\begin{figure}
    \centering
    \includegraphics[height=8cm,width=1\columnwidth]{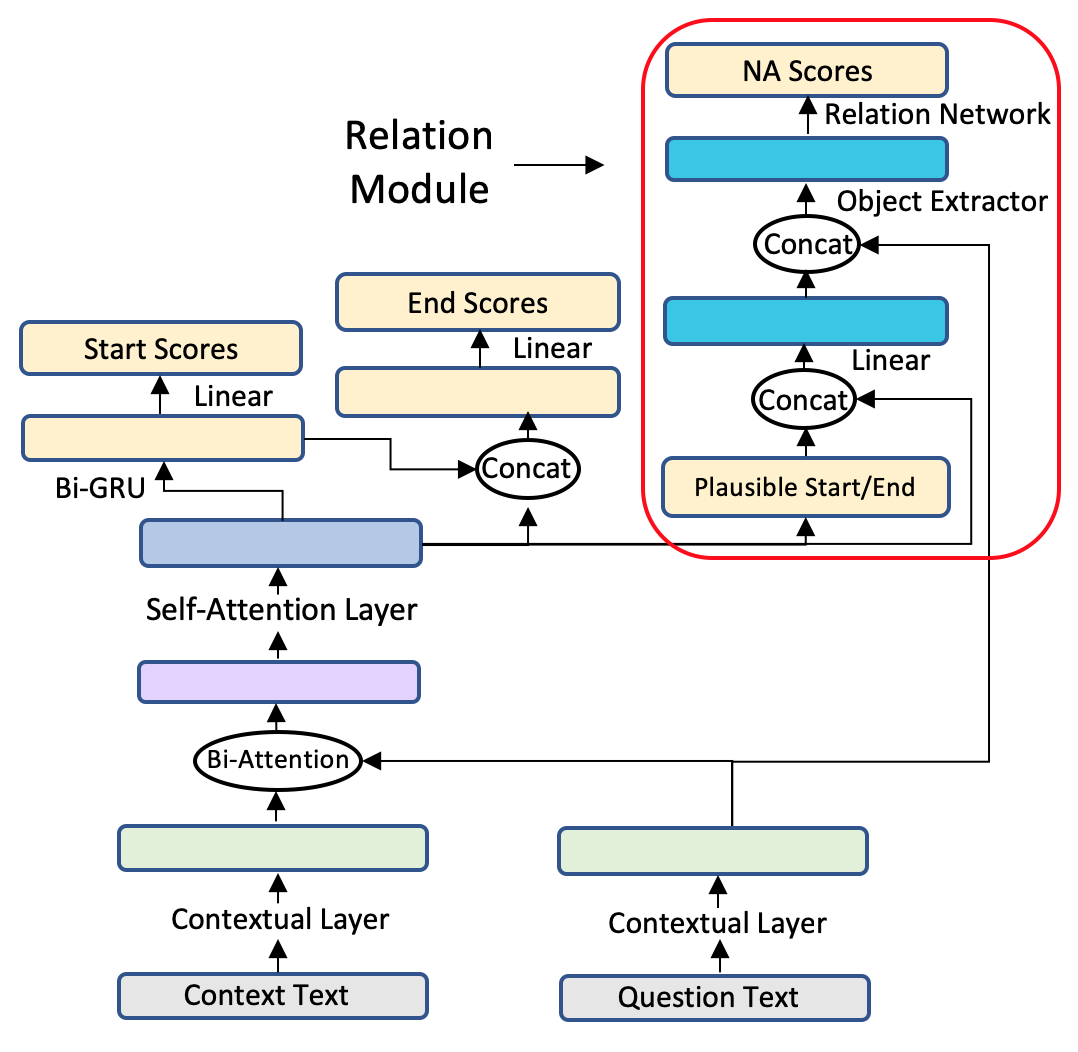}
    \caption{Relation Module applied on BiDAF.}
    \label{fig:DocQARelation}
\label{figure:DocQARelation}
\end{figure}

The input to the relation module is the context vector that is generated from the bi-directional attention flow layer. This context layer is augmented with the hidden states of linear layers trained against plausible answers, which also takes the context layer from the attention flow layer as input. This configuration is shown in Figure~\ref{figure:DocQARelation}. 
\subsection{BERT}
BERT is a masked language model pre-trained on large amounts of data that is the core component of all of the current state-of-the-art models on the SQuAD 2.0 task. 
The input to BERT is the concatenation of a question and context pair in the form of [``CLS''; question; ``SEP''; context; ``SEP'']. BERT comes with its own special ``CLS'' token, which is pre-trained on a next sentence pair objective in order to encode entailment information between the two sentences during the pre-training scheme. 

We leverage this ``CLS'' node with the relation module by concatenating it with the output of our Relation Module, and projecting the values down to a single dimension. This combines the information stored in the ``CLS'' token that has been learned from the pre-training, as well as the information that we learn through our relation module. We allow gradients to be passed through all layers of BERT, and finetune the initialized weights with the SQuAD 2.0 dataset.

\section{Experiments}
We experiment on the SQuAD 2.0 dataset \cite{RajpurkarJL18SQuAD2} which contains question and context examples that are crowd-sourced from Wikipedia. 
Each example contains an answer span in the passage, or an empty string, indicating that an answer doesn't exist. The results are reported on the SQuAD 2.0 development set. 

We use the following parameters in our BiDAF experiment: $16$ context heads, $2$ question heads. We set our regularization loss weight for the object extractor to be 0.0005. We use Adam optimizer \cite{AdamKingmaB14}, with a start learning rate of $0.0008$ and decay the learning rate by $0.5$ with a patience of $3$ epochs. We add auxiliary losses for plausible answers, and re-rank the non-answerable loss as in \cite{Hu2018ReadV}. 

BERT comes in two different sizes, a BERT-base model (comprising of roughly 110 million parameters),  and a BERT-large model (comprising of roughly 340 million parameters). We use the BERT-base model to run our experiments due to the limited computing resources that training the BERT-large model would take. We only use the BERT-large model to show that we still get improvements with the relation module. The relation module on top of the BERT-base model only contains roughly 10 million parameters.

We use the BERT-base model to run our experiments with the same hyper-parameters given on the official BERT GitHub repository. We use $16$ context objects, $2$ question heads, and a regularization loss of 0.0005. We also show that on top of the BERT-large model, on the development set, our relation module still obtains performance gain\footnote{We do not have enough time to get official SQuAD 2.0 evaluation results for the large BERT models.}. We use the same number of objects, and the same regularization losses for the BiDAF model experiments.

\begin{table}[t!]
\begin{small}
\begin{center}
\begin{tabular}{|l|r|l|}
\hline \bf Model & \bf EM(\%) & \bf F1(\%)\\ \hline
\cite{Simple_and_effective} & 61.9 & 64.8 \\ \hline
Our Implementation of BiDAF & 65.7 & 68.6 \\ 
BiDAF + Relation Module & {\bf 67.7} & {\bf 70.4} \\ \hline
BERT-base & 73.6  & 76.6 \\
BERT-base + Relation Module & {\bf 74.2} & {\bf 77.6} \\ \hline
BERT-large & 78.9 & 82.1 \\
BERT-large + Relation Module & {\bf 79.2} & {\bf 82.6} \\
\hline
\end{tabular}
\end{center}
\end{small}
\caption{\label{font-table} Model performance on SQuAD 2.0 development set averaged over three random seeds.
}
\label{table:res}
\end{table}

\begin{table}[t!]
\begin{small}
\begin{center}
\begin{tabular}{|l|r|l|}
\hline \bf Model & \bf EM(\%) & \bf F1(\%)\\ \hline
BERT-base & 70.7 & 74.4 \\
BERT-base +Answer Verifier & 71.7 & 75.5\\
BERT-base + Relation Module & \bf 73.2 & \bf 76.8 \\ 
\hline
\end{tabular}
\end{center}
\end{small}
\caption{\label{font-table} SQuAD 2.0 leaderboard numbers on the BERT-base Models. Our model shows improvement over the public BERT-base models on the official evaluation.
}
\label{table:test}
\end{table}

Table~\ref{table:res} presents the results of the baseline readers with and without the relation module on the SQuAD development set. Our proposed relation module improves the overall F1 and EM accuracy: $2.0$\% gain on EM and $1.8$\% gain on F1 on the BiDAF, as well as $0.8$\% gain on EM and $1.0$\% gain on F1 on the BERT-base model. 
Our relation module is able to take relational information between object-pairs and form a better no-answer prediction than a model without it. The module obtains less gain ($0.5\%$ gain of F1) on BERT large model due to the better performance of BERT large model. This module is reader independent and works for any reading comprehension model related to non-answerable tasks. 


Table~\ref{table:test} presents performance of three BERT-base models with minimum additions taken from the official SQuAD 2.0 leaderboard. We see that our relation module gives more gain than an Answer Verifier on top of the BERT-base model. Our module gains $1.3\%$ F1 over the Answer Verifier.

\begin{table}[t]
    \begin{small}
    \bigskip
    \begin{center}
    \begin{tabular}{|l|c|c|}
        \hline
               & Answerable & Non-Answerable \\
        \hline
        BERT-base & 81.5 & 78.3 \\
        \hline
        + Relation Module   & \bf82.1  & \bf82.1 \\
        \hline
    \end{tabular}
    \end{center}
    \end{small}
    \caption{Prediction accuracies on answerable and non-answerable questions on development set.}\label{tbl:pred_ans}
\end{table}

Since our relation module is designed to help a MRC model's ability to judge non-answerable questions, we examine the accuracy when a question is answerable and when a question is non-answerable. Table \ref{tbl:pred_ans} compares these accuracy numbers for these questions with and without the relation module on top of the BERT-base model.
The relation module improves prediction accuracy for both types of questions, and with more accuracy gain on the non-answerable questions: close to $4\%$ gain on the non-answerable questions, which is more than $200$ non-answerable questions are correctly predicted. 

\begin{table}[t!]
\begin{small}
\begin{center}
\begin{tabular}{|l|c|c|}
\hline \bf Model & \bf EM(\%) & \bf F1(\%)\\ \hline
BERT-base & 73.6 & 76.6 \\
BERT-base+Plausible Answers & 73.5 & 76.9\\
BERT-base+RM-Plausible Answers & 73.6 & 76.9 \\ 
BERT-base+RM (4 heads) &  74.1 & 77.4 \\ 
BERT-base+RM (16 heads) & \bf 74.2 & \bf 77.6 \\ 
BERT-base+RM (64 heads) & 74.0 & 77.2 \\ 
\hline
\end{tabular}
\end{center}
\end{small}
\caption{\label{font-table} Ablation study on our Relation Module. We experiment with just having plausible answers, just having relation network, and different number of heads for the objects extracted by the relation network. Each of these values are averaged over three random seeds.
}
\label{table:ablation}
\end{table}

\section{Ablation Study}
We conduct an ablation study to show how different components of the relation module affects the overall performance for the BERT-base model. First we test only adding plausible answers on top of the BERT-base model, in order to quantify the gain in span prediction that adding these extra answers in would give. We show that with just adding plausible answers, the average of the three seeds gain only about a $0.3$ F1. This gain in F1 is due to the BERT layers being fine-tuned on more answer span data that we provide. 
Next we study the effects of removing augmenting the context vector with plausible answers. We feed the output of our BERT-base model directly into the object extractor and subsequently to the relation network. This quantifies the effect of forcing the self-attentive heads to focus on a plausible answer span. We notice that this performs comparably to just adding plausible answers, also with only around a $0.3$ F1 gain.

Finally, we conduct a study to see the effects of different number of heads on our relation module. We experiment with 4, 16, and 64 heads, with $16$ heads performing the best out of these three configurations. Having too few heads hinders the performance due to not enough information being propagated for the relation network to operate on. Having too many heads will introduce redundant information, as well as incorporating extraneous noise for our model to sift through to generate meaningful relations.

\section{Analysis}
In order to gain better understanding on how the relation module helps on the unanswerable prediction, we examine the objects extracted from the multi-head self-attentive pooling. This is to check whether the relevant semantics are extracted for the relation network. Examples are selected from the development set for data analysis. 
\begin{figure}[tp]
\framebox{
\parbox{0.45\textwidth}{
\small
\textbf{\textcolor{blue}{Example 1} } \newline
\textbf{Context:}  Each year, the southern California area has about 10,000 earthquakes. Nearly all of them are so small that they are not felt. Only several hundred are greater than magnitude 3.0, and only about 15–20 are greater than magnitude 4.0. The magnitude 6.7 1994 \textbf{\textcolor{red}{Northridge earthquake}} was particularly destructive, causing a substantial number of deaths, injuries, and structural collapses. It caused the most property damage of any earthquake in U.S. history, estimated at over \$20 billion.\newline
\textbf{Question:} What earthquake caused \$20 million in damage? \newline
\textbf{Answer:} None.
}
}
\label{fig:exp1}
\end{figure}

In Example $1$, the BERT-base model incorrectly outputs ``Northridge earthquake'' (in red) as the answer. However, after adding our relation module, the model rejects this possible answer and outputs a non-answerable prediction. 
\begin{figure}[t]
  \includegraphics[width=1.0\linewidth]{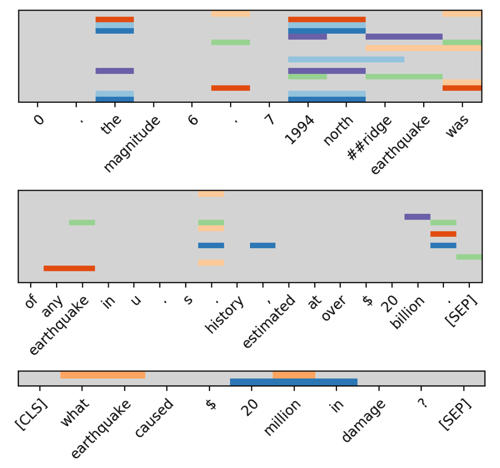}
  \caption{In each subplot, each row represents one object from our object extractor; for each object we highlight the top 5 tokens with highest weights in the entire context and question. We show the two windows where the majority of these top 5 weights occur. For example, the top purple object in the context looks at key phrases such as ``\#\#ridge earthquake'' in the top subplot and ``billion'' in the middle subplot; the blue object in the question looks at ``20 million in'' in the bottom subplot.}
  \label{fig:obj1}
\end{figure}

The two objects from the question highly attend to token ``million'' (see the bottom  subplot in Figure~\ref{fig:obj1}). The top row purple object covers token ``1994'' , ``\#\#ridge earthquake'' in the possible answer span window, and ``billion'' near the end of the context window. We hypothesize that the relation network rejects the possible answer ``Northridge earthquake'' due to  the mismatch of ``million'' in the question objects and ``billion'' in the purple context object, and relation scores from all other object pairs.

\begin{figure}[thp]
\framebox{
\parbox{0.45\textwidth}{
\small
\textbf{\textcolor{blue}{Example 2} } \newline
\textbf{Context:} Even though some proofs of complexity-theoretic theorems regularly assume some concrete choice of \textbf{\textcolor{red}{input encoding}}, one tries to keep the discussion abstract enough to be independent of the choice of encoding. This can be achieved by ensuring that different representations can be transformed into each other efficiently. \newline
\textbf{Question:} What is the abstract choice typically assumed by most complexity-theoretic theorems? \newline
\textbf{Answer:} None.
}
}
\label{fig:exp2}
\end{figure}

\begin{figure}[h]
 \centering
  \centering
  \includegraphics[width=1\linewidth]{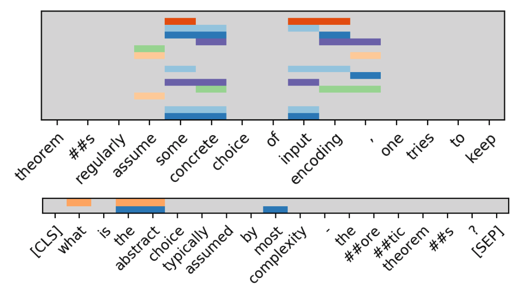}
  \caption{In each subplot, each row represents one object from our object extractor; for each object we highlight the top 5 tokens with highest weights in the entire context and question. We show a window where the majority of the top 5 weights occur. For example, there are numerous objects in the context window that look at the key phrase ``some concrete'' in the top subplot; the two objects in the question look at the key phrase ``the abstract'' in the bottom subplot.}
  \label{fig:obj2}
\end{figure}

Example $2$ shows another example of non-answerable question and context pair.  The BERT-base model incorrectly outputs ``input encoding'' (in red) as its prediction, while adding our relation module on the BERT-base model predicts correctly that the question is not answerable. Figure~\ref{fig:obj2} gives a visual illustration of objects extracted from context and question. In Figure~\ref{fig:obj2}, the upper plot illustrates the $16$ semantic objects shown in this context window and the lower plot illustrates the two semantic objects from the question. 
We see that from the upper plot, ``some concrete'' and ``input encoding'' are highlighted, while in the lower plot, ``what'', ``the abstract'', ``most'' are highlighted. The mismatch of ``the abstract'' from the question objects and ``some concrete'' from the context objects helps indicate that the question is unanswerable.

\section{Conclusion}
 
In this work we propose a new relation module that can be applied on any MRC reader and help increase the prediction accuracy on non-answerable questions. We extract high level semantics from multi-head self-attentive pooling. The semantic object pairs are fed into the relation network which makes a guided decision as to whether a question is answerable. In addition we augment the context vector with plausible answers, allowing us to extract objects focused on the proposed answer span, and differentiate from other objects that are not as relevant in the context.
Our results on the SQuAD 2.0 dataset using the relation module on both BiDAF and BERT models show improvements from the relation module.
These results prove the effectiveness of our relation module.

For future work, we plan to generalize the relation module to other aspects of question answering, including span prediction or multi-hop reasoning. 

\section{Acknowledgements}
We would like to thank Robin Jia and Pranav Rajpurkar for running the SQuAD evaluation on our submitted models.

\bibliography{main}
\bibliographystyle{acl_natbib}

\appendix

\end{document}